\documentclass[pdflatex,sn-mathphys-num]{sn-jnl}

\usepackage{graphicx}%
\usepackage{multirow}%
\usepackage{amsmath,amssymb,amsfonts}%
\usepackage{amsthm}%
\usepackage{mathrsfs}%
\usepackage[title]{appendix}%
\usepackage{xcolor}%
\usepackage{textcomp}%
\usepackage{manyfoot}%
\usepackage{booktabs}%
\usepackage{algorithm}%
\usepackage{algorithmicx}%
\usepackage{algpseudocode}%
\usepackage{listings}%

\theoremstyle{thmstyleone}%
\theoremstyle{thmstyletwo}%

\theoremstyle{thmstylethree}%

\begin{document}

\title[Article Title]{\textbf{ Sustainable Deep Learning-Based Breast Lesion Segmentation: Impact of Breast Region Segmentation on Performance}}

\author*[ 1,2,3]{\fnm{Sam} \sur{Narimani}}\email{sam.narimania@gmail.com }

\author[ 3,4]{\fnm{Solveig} \sur{Roth Hoff}}\email{solveig.roth.hoff@ntnu.no}


\author[ 5,6]{\fnm{Kathinka} \sur{Dæhli Kurz}}\email{kathinka.dehli.kurz@sus.no}

\author[ 7]{\fnm{Kjell-Inge} \sur{Gjesdal}}\email{k.i.gjesdal@ioks.uio.no}

\author[ 8,9]{\fnm{Jürgen} \sur{Geisler}}\email{jurgen.geisler@medisin.uio.no}

\author[ 1,2,3]{\fnm{Endre} \sur{Grøvik}}\email{endre.grovik@gmail.com}

\affil[1]{\small{\orgdiv{ Department of Physics}, \orgname{Norwegian University of Science and Technology}, \orgaddress{ \city{Trondheim},  \country{Norway}}}}

\affil[2]{\small{\orgdiv{ Research and Development Department}, \orgname{More og Romsdal Hospital Trust}, \orgaddress{ \city{Aalesund},  \country{Norway}}}}

\affil[3]{\small{\orgdiv{ Department of Radiology}, \orgname{More og Romsdal Hospital Trust}, \orgaddress{ \city{Aalesund},  \country{Norway}}}}

\affil[4]{\small{\orgdiv{ Department of Health Sciences}, \orgname{Norwegian University of Science and Technology}, \orgaddress{ \city{Aalesund},  \country{Norway}}}}

\affil[5]  {\small {\orgdiv{ Stavanger Medical Imaging Group, Radiology Department}, \orgname{Stavanger University Hospital}, \orgaddress{ \city{Stavanger},  \country{Norway}}}}

\affil[6]  {\small {\orgdiv{ Department of Electrical Engineering and Computer Science}, \orgname{The University of Stavanger}, \orgaddress{ \city{Stavanger},  \country{Norway}}}}

\affil[7] {\small  \orgdiv{ Department of Diagnostic Imaging}, \orgname{Akershus University Hospital}, \orgaddress{ \city{Lorenskog},  \country{Norway}}}

\affil[8] {\small  \orgdiv{ Institute of Clinical Medicine}, \orgname{University of Oslo}, \orgaddress{ \city{Lorenskog},  \country{Norway}}}

\affil[9] {\small  \orgdiv{ Department of Oncology}, \orgname{Akershus University Hospital}, \orgaddress{ \city{Lorenskog},  \country{Norway}}}




\abstract{\textbf{Purpose:} Segmentation of the breast lesion in dynamic contrast-enhanced magnetic resonance imaging (DCE-MRI) is an essential step to accurately diagnose and plan treatment and monitor progress. This study aims to highlight the impact of breast region segmentation (BRS) on deep learning-based breast lesion segmentation (BLS) in breast DCE-MRI.

\textbf{Methods:} Using the Stavanger Dataset containing primarily 59 DCE-MRI scans and UNet++ as deep learning models, four different process were conducted to compare effect of BRS on BLS. These four approaches included the whole volume without BRS and with BRS, BRS with the selected lesion slices and lastly optimal volume with BRS. Preprocessing methods like augmentation and oversampling were used to enhance the small dataset, data shape uniformity and improve model performance. Optimal volume size were investigated by a precise process to ensure that all lesions existed in slices. To evaluate the model, a hybrid loss function including dice, focal and cross entropy along with 5-fold cross validation method were used and lastly a test dataset which was randomly split used to evaluate the model performance on unseen data for each of four mentioned approaches.

\textbf{Results:} Results demonstrate that using BRS considerably improved model performance and validation. Significant improvement in last approach- optimal volume with BRS- compared to the approach without BRS counting around 50 percent demonstrating how effective BRS  has been in BLS. Moreover, huge improvement in energy consumption, decreasing up to 450 percent, introduces a green solution toward a more environmentally sustainable approach for future work on large dataset.
 }




\maketitle

\section{Introduction}\label{sec1}

Breast cancer is the most common cancer in women and the second leading cause of cancer death in women worldwide \cite{mcpherson2000breast}. Although number of deaths due to breast cancer has slightly decreased over the past 30 years in Norway thanks to various screening programs, 619 women lost their lives to breast cancer in 2022 \cite{cancerinnorway2023,brystkreft2024}. Therefore, continuous research on new methods to improve the detection and characterization of breast cancer is vital to mitigate death due to breast cancer.

Medical imaging is crucial for breast cancer diagnosis, and imaging modalities used include mammography, ultrasound, and MRI. Dynamic contrast enhanced magnetic resonance imaging (DCE-MRI) is the most sensitive technique and is mainly used for staging of known breast cancer, evaluation of response to neoadjuvant chemotherapy and screening of women with increased risk \cite{mann2019breast}.  Breast DCE-MRI is a multiparametric technique including T1, T2 and diffusion weighted imaging. It is performed with intravenous injection of a gadolinium chelate that shortens the T1 time and leads to higher signal on T1 weighted images. Malignant breast tumors generally have more permeable vessels than benign breast tissue, with faster extravasation of contrast agent and rapid enhancement \cite{knopp1999pathophysiologic}. Malignant breast tumors either present as contrast enhancing solid masses or as pathological non-mass enhancement.  Malignant findings must be separated from benign contrast enhancing masses and from normal background parenchymal enhancement, based either on contrast kinetics, characteristics like lesion shape, margins and internal enhancement characteristics or by biopsy (BI-RADS).

Artificial intelligence demonstrates an increasing potential in diagnostic and breast cancer detection \cite{WHO2020,alshawwa2024advancements}. AI algorithms can analyze breast DCE-MRI and is able to acceptably detect, segment and classify abnormalities in breast anatomy \cite{yue2022deep,benjelloun2018automated,zhang2019deep}. Among all AI tasks, segmentation plays a significant role in breast cancer detection and characterization since it localizes lesions and reduces ambiguity by isolating region of interest (ROI), allowing to conduct quantitative analysis on ROI \cite{zhang2022automatic,patra2021breast}. However, there are also several challenges with segmentation of breast cancer tumors such as imbalance class due to small lesions \cite{vidal2022u}. Numerous research projects in breast cancer segmentation were carried out by utilizing Machine Learning (ML) and Deep Learning (DL) models \cite{dhungel2015deep,patel2010adaptive,dominguez2009toward}. Convolutional Neural Network (CNN) based DL models have gained increasing attraction and as a result one of the most common DL models, UNet, was introduced in 2015 \cite{ronneberger2015u}. This model has paved the way for the next generation of variations such as Connected-UNet \cite{baccouche2021connected} and UNet++ \cite{zhou2018unet++}.

While numerous well-known and prominent DL models have been confirmed to segment in breast DCE-MRI, there has not been any studies focusing on how the whole breast region can affect breast lesion segmentation performance. Furthermore, the DL models in use are associated with high computational costs, which represents a negative environmental aspect. In our study we aim to explore whether breast region segmentation (BRS) can enhance breast lesion segmentation (BLS) performance and also if the use of BRS can decrease the training time and the carbon footprint.

\section{Methodology}\label{sec2}

\subsection{Data Pre-processing and insight}\label{subsec2}

\subsubsection{Data Pre-processing}\label{subsec2}

\begin{figure}[h!]
\centering
\includegraphics[width=1\textwidth, trim=0 0 0 0, clip]{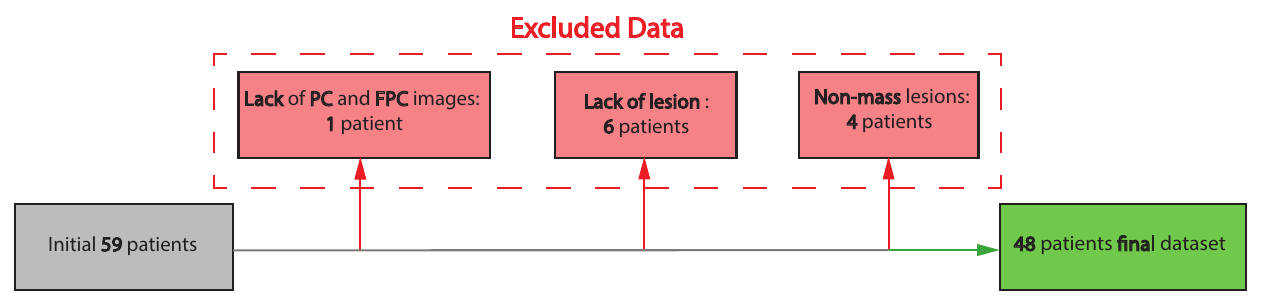}
\caption{\centering Process of data exclusion based on available information in Stavanger cohort (Pre-Contrast (PC) and First Post-Contrast (FPC)). Red and green lines indicate exclusion and inclusion of data,respectively)} \label{CleanData}
\end{figure}

The dataset employed in this study, referred to as the Stavanger dataset, was sourced from Stavanger University Hospital. A detailed description of the dataset's characteristics and features has been provided in our previous work \cite{narimani2024comparative}. The dataset includes sequences from breast DCE-MRI, specifically pre-contrast and first post-contrast images collected for this study.
Originally, the dataset contained data from 59 patients; however, 11 patients were excluded due to missing or incomplete information. 
As a result, the final cohort consisted of 48 patients with complete and relevant data for further analysis. The stepwise exclusion process is summarized in Figure \ref{CleanData}.

To standardize the data preparation pipeline, the pre-contrast and first post-contrast images were automatically identified for each patient and subsequently converted from the DICOM format, the standard in medical imaging, to the NIFTI format, which is used in our models. The pipeline also incorporated a mechanism to identify missing data, such as cases without pre-contrast or post-contrast images. To ensure consistent volumetric representation throughout the analysis, random oversampling was implemented, enhancing not only the uniformity of the dataset but also simplifying the data-loading process for the model. To further ensure data uniformity, all images were reoriented to the standard RAS orientation commonly utilized in breast imaging within Scandinavian countries.

\begin{figure}[h!]
\centering
\includegraphics[width=1.1\textwidth, trim=15 40 0 10, clip]{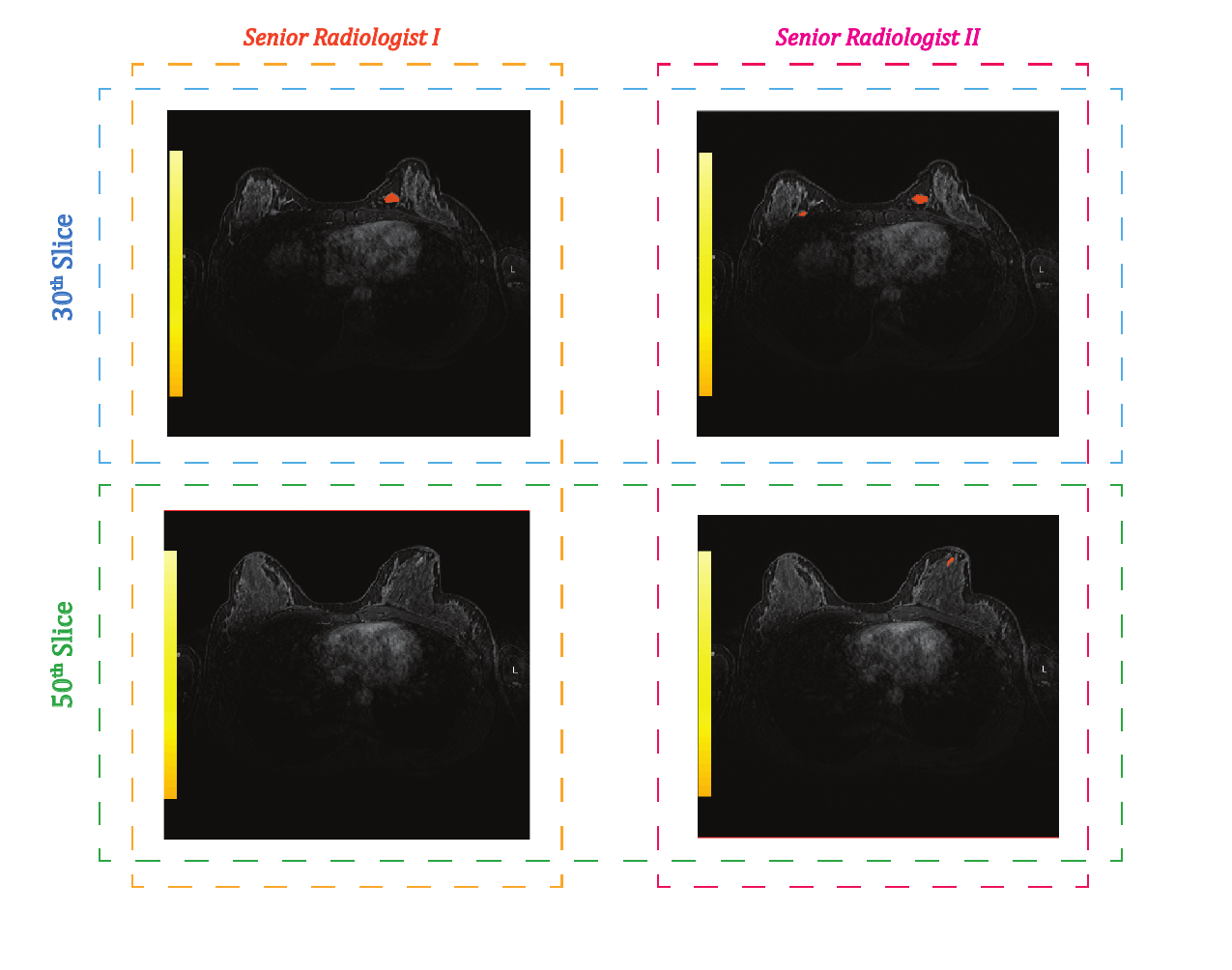}
\caption{\centering Annotation comparison between two senior radiologists on a patient in test dataset } \label{Annotation}
\end{figure}

Additionally, subtraction images and input types (pre-contrast or post-contrast images) were integrated into the pipeline as required. Regarding the mask files, all lesion annotations were performed by an experienced senior breast radiologists with extensive expertise in breast cancer diagnostics. For the training datasets, only the largest lesions were annotated. In contrast, the test datasets were annotated by another senior breast radiologist to ensure that all lesions were considered in the evaluation metrics. 
Figure \ref{Annotation} illustrates the annotation process for a patient in test dataset conducted by two senior breast radiologists. As shown in the figure, Radiologist 1 focused only on the largest lesions and was therefore excluded from the test evaluation. In contrast, Radiologist 2 annotated all detectable lesions, ensuring a comprehensive evaluation of the test dataset.

\subsubsection{Data Analysis and Insights}\label{subsubsec2}

\begin{figure}[h!]
\centering
\includegraphics[width=1.1\textwidth, trim=15 5 0 0, clip]{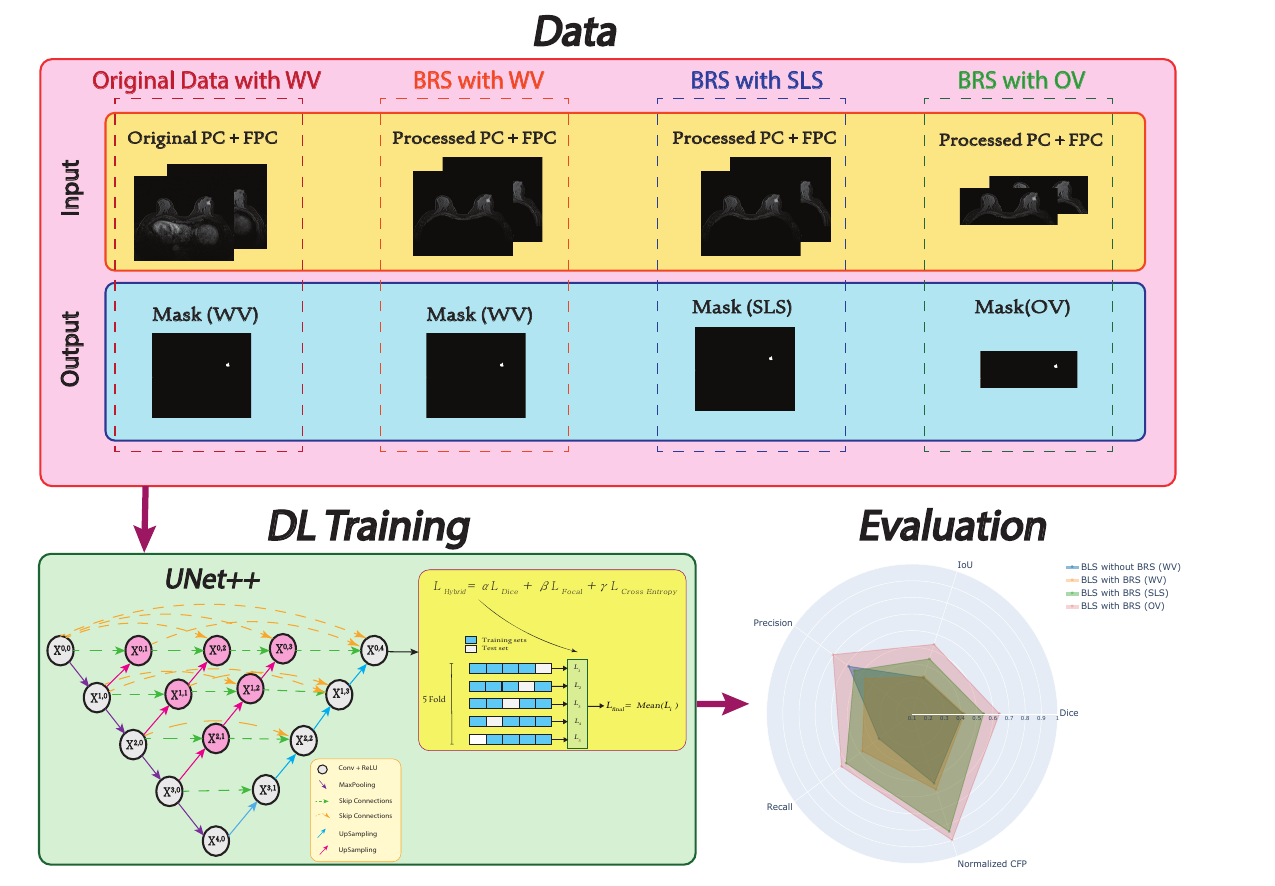}
\caption{\centering Overview of data configurations (WV without BRS, WV with BRS, SLS with BRS, and OV with BRS), the UNet++ model, a hybrid loss function (Dice, Focal, and Cross-Entropy), and evaluation metrics (IoU, Dice, Precision, Recall, and Normalized CFP)} \label{PaperDesign}
\end{figure}

This study primarily concentrates on effect of BRS on BLS. To investigate this, two separate groups of BLS approaches were considered: one with BRS applied to all images and the other without BRS. However, this problem could be explored further using a more sustainable and environmentally friendly approach by incorporating additional data analysis steps. Therefore, after applying BRS to both pre- and post-contrast images, the study explores three potential volume strategies: the Whole Volume (WV), the Selected Lesion Slices (SLS)—which include only the slices containing lesions— and The Optimized Volume (OV), which incorporates 2D slice optimization in SLS. Consequently, four different datasets including original data with WV, BRS with WV, BRS with SLS, and BRS with OV were created for this study. These four types of datasets, utilized separately in deep learning (DL) models, are illustrated in Figure \ref{PaperDesign} in the Data section. Notably, the shape of the label files corresponds directly to the structure of their respective input datasets.

To further analyze the steps leading to approaches with BRS, a pretrained model previously developed was utilized for whole breast segmentation using BRS. By applying the mask images predicted by BRS to corresponding images, new images are created. These new images exclude noise from low-intensity areas anteriorly and remove organs such as the heart and lungs posteriorly. This process isolates the breast anatomy, enabling a clearer analysis of its internal structure. By focusing solely on the breast, it becomes easier to optimize the sites where lesions are most likely to exist.
 
To identify the optimized height, data analysis was performed on all images to ensure that all lesions fall within the selected region. For each patient, the first pixel with a non-zero value was detected from the top and bottom of the slices. The maximum and minimum distances from coordinates across all slices for each patient were then identified. Similarly, the maximum depth of the breast along the chest wall was determined to ensure the image can be cropped with a safe margin, guaranteeing that all lesions remain within the cropped region. Based on these calculations, the maximum height of the selected area across all patients was chosen and applied uniformly.

Additionally, deep learning (DL) model and network configuration considerations were addressed to ensure compatibility with the new image shape. The input dimensions were adjusted to be a multiple of 32, meeting the requirements of the current DL model and ensuring compatibility with other well-known segmentation models for future work.

\begin{figure}[h!]
\centering
\includegraphics[width=1\textwidth, trim=30 185 50 50, clip]{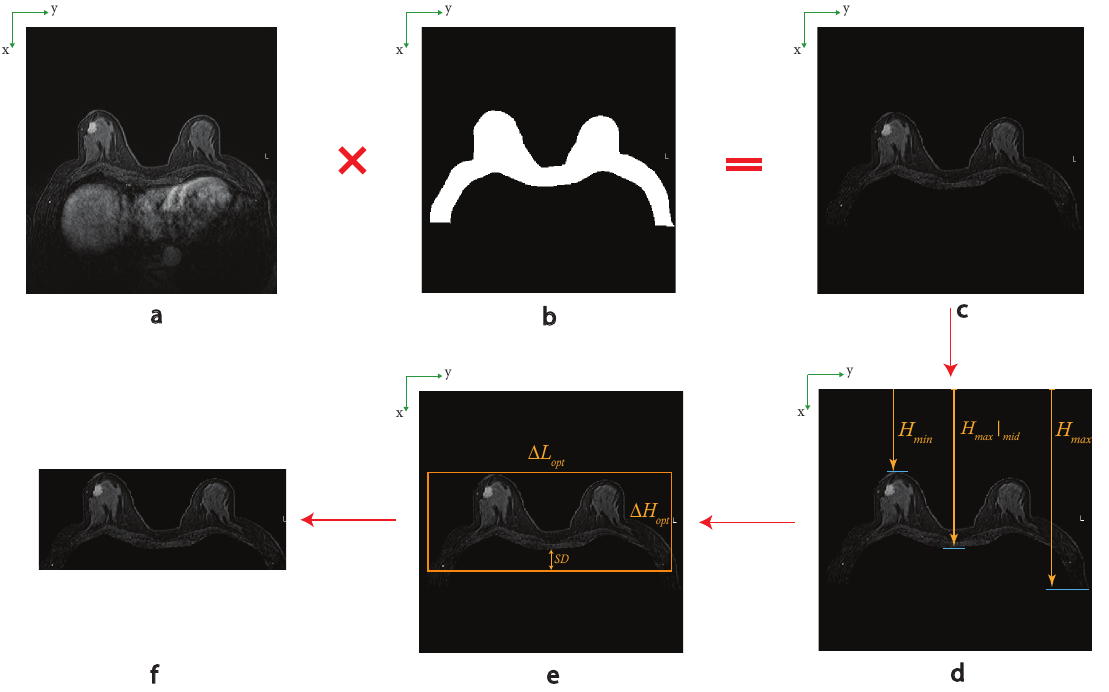}
\caption{\centering Process of image size optimization (a : original pre and first post contrast images, b: mask file of corresponding patient predicted by pre-trained BRS model , c: new images by multiplying original and corresponding predicted mask file , d: finding the maximum and minimum of non-zero pixels in image and middle of breast chest , e: optimizing the rectangle containing the region of interest (SD stands for Safe Distance) , f: final slice for training of OV) } \label{DataAnalysis}
\end{figure}

Figure \ref{DataAnalysis} illustrates the process of refining the images to create a smaller, more focused region of interest, enhancing the efficiency and precision of subsequent analyses.

\subsection{Deep learning network}\label{subsec2}

Segmentation DL models, such as UNet and UNet++, are widely used for segmenting small objects like lesions, primarily using Convolutional Neural Networks (CNNs) \cite{hesaraki2024unet++,kanadath2024air,robin2021breast}. These models are based on a contraction and expansion process, commonly referred to as the encoder and decoder, respectively. The encoder is responsible for extracting and identifying the most relevant features at each encoding level, while the decoder reconstructs these compressed features to produce the desired output. Employed skip connections between encoder and decoder enables the model to remember the features that have been forgotten in decoder process and as a result improves model performance and segmentation accuracy \cite{zhou2018unet++,ronneberger2015u}. 

Among encoder-decoder architectures with skip connections, UNet++ has gained significant attention due to its nested skip connections and more complex structure, which enable it to establish better relationships between inputs and outputs \cite{jalalian2025improving,li2023eres}. The nested skip connections in UNet++ reduce the loss of critical features transferred between the encoder and decoder, thereby enhancing model performance and segmentation accuracy \cite{zhou2018unet++}. UNet++ architecture was illustrated in DL training section of Figure \ref{PaperDesign}. In addition, Table \ref{table:UNet++ specification} provides detailed specifications of the UNet++ architecture, including key aspects such as the number of learnable parameters, network depth, types of layers, and other distinctive features.

\begin{table}[h!]
\centering
\caption{\centering UNet++ Model Specifications}\label{table:UNet++ specification}%
\begin{tabular}{@{}lll@{}}
\hline
\textbf{Category}             & \textbf{Feature}             & \textbf{Details} \\ \hline
                              & Number of Parameters        & 2410468 \\
\textbf{General Information}  & Input Shape                 & (H, W, C) \\
                              & Output Shape                & (H, W, 1) \\
\hline
                            & Depth of Network            & 5 levels \\
\textbf{Network Architecture}       & Number of Layers    & 240 \\
                              & Base Filter Size            & 32 \\
\hline
                                & Convolution Type            & 3x3 Convolutions with stride 1 \\
\textbf{Layer Types}            & Pooling Layers              & MaxPooling 2x2 \\
                              & Upsampling Layers           & Transposed Convolution \\
\hline
                                & Activation Function         & ReLU for hidden layers, Sigmoid for output \\
\textbf{Activation and Regularization} & Normalization               & Batch Normalization \\
                                       & Dropout                     & 0.00 \\
\hline
\end{tabular}
\label{tab:unetpp-specification}
\end{table}

\subsection{Evaluation and setup}\label{subsec2}
To evaluate the performance of DL model in training, a hybrid loss function alongside 5-fold  cross validation were used. Equation \ref{equation1} demonstrates hybrid loss function employed in training process.

\begin{equation}
\begin{aligned}
\mathcal{L}_{\text{Hybrid}} &= \alpha \cdot \mathcal{L}_{\text{Dice}} + \beta \cdot \mathcal{L}_{\text{Focal}} + \gamma \cdot \mathcal{L}_{\text{Cross-Entropy}}              
\end{aligned}
\label{equation1}
\end{equation}\\

Where $\alpha$, $\beta$, and $\gamma$ are coefficients that determine the contribution of Dice loss, Focal loss, and Cross-Entropy loss, respectively, and $\alpha + \beta + \gamma = 1$ . These coefficients can be empirically tuned based on the specific requirements of the task, such as addressing class imbalance (by using higher weight for Focal Loss) or improving segmentation quality (by enhancing Dice Loss coefficient). In this study $\alpha$, $\beta$, and $\gamma$ were chosen 0.1 , 0.45 and 0.45, respectively. Dice loss \cite{milletari2016v}, Focal loss \cite{lin2017focal}, and Cross-Entropy loss \cite{mao2023cross} are additionally  defined as relations \ref{equation2}, \ref{equation3} and \ref{equation4}, respectively.

\begin{equation}
\mathcal{L}_{\text{Dice}}(P, G) = 1 - 2 \cdot \frac{|P \cap G|}{|P| + |G|}
\label{equation2}
\end{equation}

\begin{equation}
\mathcal{L}_{\text{Focal}}(p_t) = - \sum \alpha_t \cdot (1 - p_t)^{\gamma_f} \cdot \log(p_t)
\label{equation3}
\end{equation}

\begin{equation}
\mathcal{L}_{\text{Cross-Entropy}}(P, G) = - \sum G \cdot \log(P)
\label{equation4}
\end{equation}\\

Dice loss measures the overlap between the predicted segmentation \(P\) and the ground truth \(G\) \cite{milletari2016v}. On the other hand, Focal loss addresses class imbalance by reducing the weight of well-classified examples. In relation \ref{equation3}, \(p_t\) represents the predicted probability for each true class, \(\alpha_t\) is a balancing factor, and \(\gamma_f\) is a focusing parameter that adjusts the importance of harder-to-classify samples \cite{lin2017focal}. Cross-Entropy loss is a widely used loss function in classification problems which calculates the difference between the predicted probability distribution \(P\) and the true distribution \(G\). This loss penalizes incorrect predictions more heavily, ensuring that the predicted probabilities align closely with the ground truth \cite{mao2023cross}. This hybrid loss function and 5-fold cross validation was depicted in DL training part of the Figure \ref{PaperDesign}.

On the other hand to evaluate results on test dataset different metrics such as Dice, IoU, Precision and Recall can be used which are defined as relations  \ref{equation Dice} to \ref{equation Recall}.

\begin{equation}
\text{Dice} = \frac{2TP}{ 2TP + FN + FP}\label{equation Dice}
\end{equation}

\begin{equation}
\text{IoU} = \frac{TP}{ TP + FN + FP}\label{equation IoU}
\end{equation}

\begin{equation}
\text{Precision} = \frac{TP}{TP + FP}\label{equation Precision}
\end{equation}

\begin{equation}
\text{Recall} = \frac{TP}{TP + FN}\label{equation Recall}
\end{equation}

Where \(TP\), \(FP\), and \(FN\) are true positives, false positives, and false negatives, respectively.
In addition, the carbon footprint is an important factor in DL applications that should be taken into account \cite{strubell2020energy}. The production of 1 kWh of energy has an average carbon footprint of 475 grCO2 \cite{iea2019}. Therefore, the carbon footprint for each fold can be determined using Equation \ref{equationCFP}:

\begin{equation}
\text{CFP} = \frac{0.475 \cdot \text{TT}}{3600} \label{equationCFP}
\end{equation}

Where \(\text{CFP}\) and \(\text{TT}\) represents the carbon footprint in kilograms of CO2 for each fold and the training time in seconds , respectively. To better visualize the effect of CFP, the normalized CFP is defined as shown in Equation \ref{equation normcfp}:

\begin{equation}
\text{Norm}_{\text{CFP}} = 1 - (\text{CFP}_{\max} - \text{CFP}_{\min}) \left(\frac{\text{CFP}}{\text{CFP}_{\max}}\right) \label{equation normcfp}
\end{equation}

Where \(\text{Norm}_{\text{CFP}}\) indicates that a higher value corresponds to a better approach.
Setup used in this study \cite{narimani2024comparative} are the same as previous study except upgrading RAM to 64 GB and therefore, the total energy usage is approximately identical to the previous study 1 kWh for each DL training session.

\section{Results}\label{sec3}
\subsection{Experiments } \label{sec3}
\begin{figure}[h!]
\centering
\includegraphics[width=1\textwidth, trim=0 5 150 0 , clip]{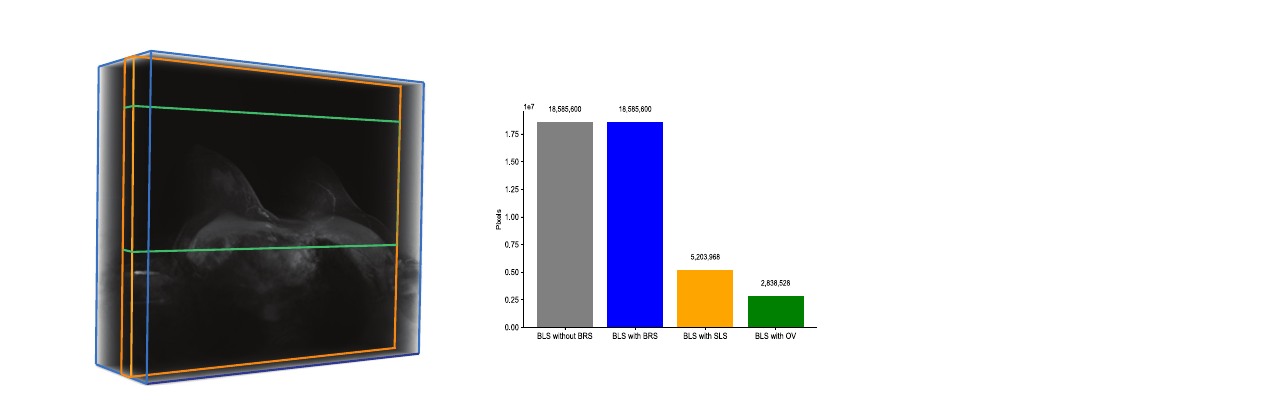}
\caption{\centering  Schematic representation of the utilized image region for each approach, including the total number of pixels analyzed per patient  } \label{Datashape}
\end{figure}

The input data compromised pre- and first-post contrast images with corresponding masks as baseline outputs. As described in data analysis section the input data and consequently masks have different shapes in each approach. Table \ref{InputShape} summarizes the final NIFTI shapes for each patient across all approaches during the training process. As shown in this table, the BRS with OV approach contains nearly 70 percent fewer slices compared to methods using the entire volume. The training process was carried out on a slice-by-slice basis, with images being provided to the model one slice at a time, as the approach employed a two-dimensional framework. Figure \ref{Datashape} additionally provides a schematic representation of the utilized image regions in each approach, along with the total number of pixels analyzed per patient.

\begin{table}[h!]
\centering
\caption{\centering Comparison of NIFTI file shape for each patient  }
\label{InputShape}
\begin{tabular}{lcccc}
\hline
\textbf{Approach} & Original Data with WV & BRS with WV & BRS with SLS & BRS with OV \\ \hline
\textbf{Input Shape} & (352,352,150) & (352,352,150) & (352,352,\textbf{42}) & (352,\textbf{192},\textbf{42}) \\ \hline
\end{tabular}
\end{table}

Training process involved UNet++ and using 5-fold cross validation along with hybrid loss function. To minimize the loss function during each epoch, RAdam optimizer with an initial learning rate of 0.001  in continuity with a ReduceLROnPlateau scheduler was utilized to boost convergency and performance of model. The scheduler adaptively modified the learning rate according to validation performance metrics, with the goal of reducing the hybrid loss function while improving training efficiency. Across all approaches, batch size of 8 was utilized with data shuffle just for training. Lastly, a random subset including two patients were split to test the model performance on previously  unobserved data. Other hyperparameters remained unchanged during all approaches to precisely analyze other effective parameters.

\subsection{Effect of BRS and Data Analysis}\label{sec3}
\begin{figure}[h!]
\centering
\includegraphics[width=1.05\textwidth, trim=10 7 0 0 , clip]{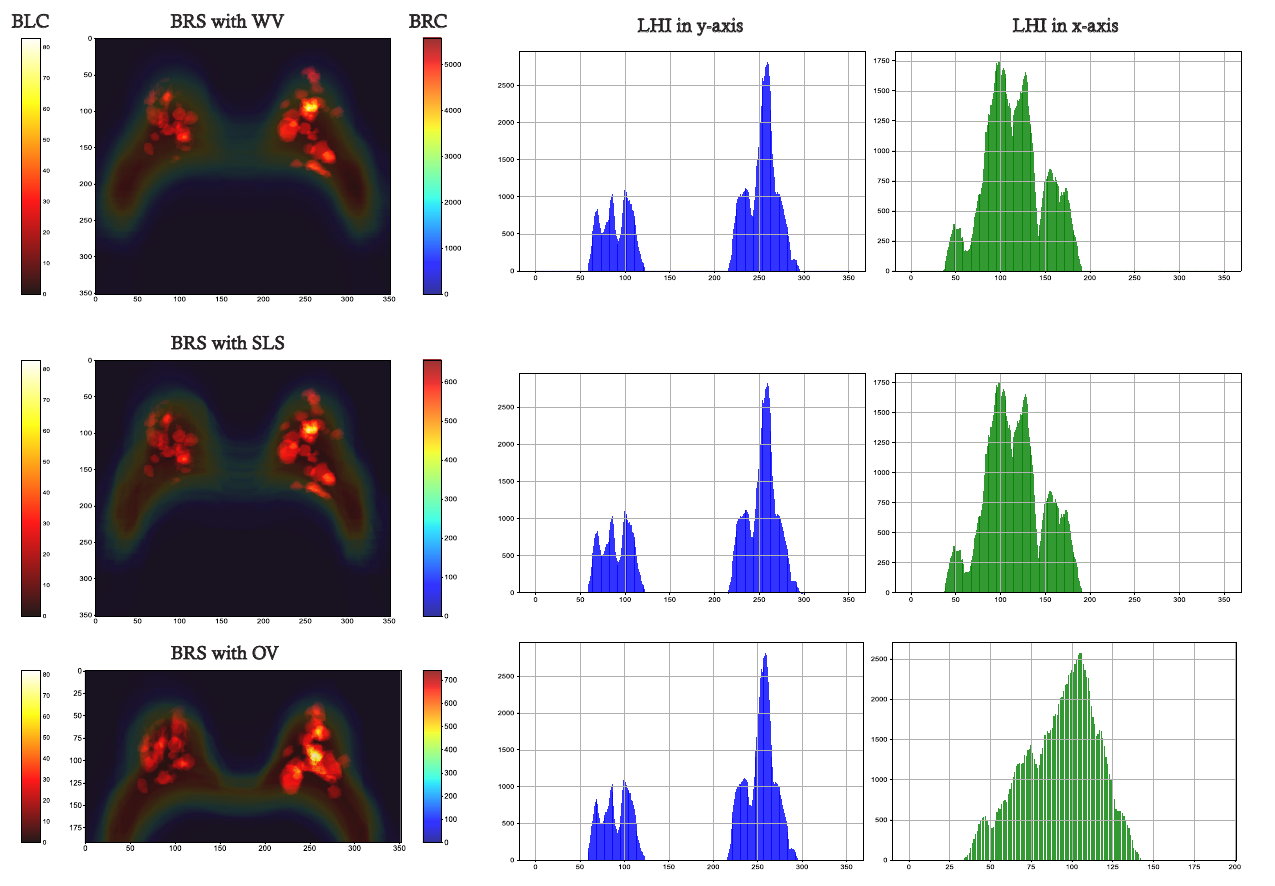}
\caption{\centering  Overlay map of total lesion masks on total breast region masks ( BLC, BRC and LHI represent Breast Lesion Colorbar, Breast Region Colorbar and Lesion Histogram Intensity, respectively)  } \label{OverlayMap}
\end{figure}

To investigate the role of BRS and data analysis on the results, we focus on the last three datasets illustrated in Data section of Figure \ref{PaperDesign}, which utilize BRS-predicted masks. By summing all breast region masks into a single slice and applying the same process for lesion masks, we can overlay the lesion total map onto the breast region total map. This approach provides a clearer visualization of how the breast region and lesion masks are distributed across the image slices. Figure \ref{OverlayMap} illustrates the overlay maps for three approaches: BRS with WV, BRS with SLS, and BRS with OV, shown from top to bottom. Alongside these maps, lesion histograms are presented in both the x- and y-directions. The overlay maps for WV and SLS demonstrate identical lesion distributions, as the lesion masks are similar. However, the presentation of the breast region masks differs significantly, as indicated by their respective color-bar ranges. A closer inspection of the SLS overlay map reveals a noticeable difference in the body midline, where fewer slices are included compared to the WV approach. In contrast, the OV overlay map shows an improved distribution of the breast region and lesion masks, where both are more concentrated in their respective areas. This suggests that the breast region is distributed across a smaller spatial area, while the lesions are more localized in both breasts.

\begin{figure}[h!]
\centering
\includegraphics[width=1.05\textwidth, trim=20 10 0 0, clip]{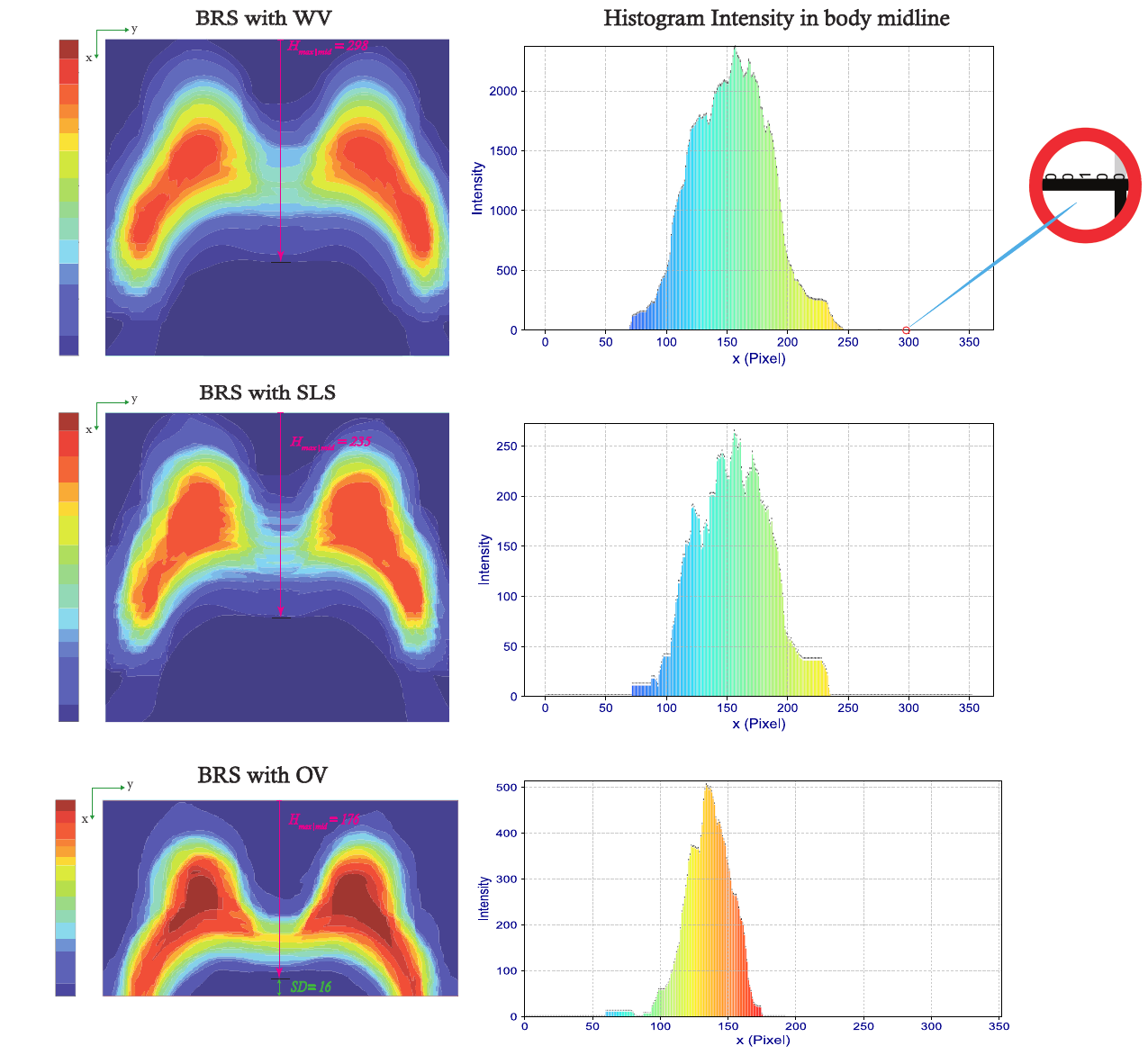}
\caption{\centering  Breast region map for all patients in the middle line parallel to x-axis } \label{BRS16Color}
\end{figure}

The lesion histogram along the y-direction reveals that lesions are predominantly concentrated in the left breast. Meanwhile, in the x-direction, the lesion distribution decreases in the OV approach compared to WV and SLS. Specifically, lesions in WV and SLS approaches appear around x = 190 from the origin, while in the OV approach, they shift to approximately x = 140. This indicates a reduced SD in lesion distribution in the OV approach compared to the other two approaches.
Finally, it is important to investigate distribution of breast region in the body midline and parallel to x-axis. This can in fact introduce a wider range for \( H_{\text{max} \vert \text{mid}} \) 
 compare to one in lesion mask distribution. Figure \ref{BRS16Color} illustrates the distribution of breast region presence in the middle line. As seen in the figure, WV shown a largest \( H_{\text{max} \vert \text{mid}} \)  by amount of 298 and the second, SLS approach, with 235 and the minimum belong to OV approach with just 176. By considering the model configurations need and 176 for OV, size of 192 was chosen for OV meaning that SD was chosen 16 pixels or approximately 1.6 cm from middle of the breast chest.

\subsection{Model Performance and Generalization}\label{sec3}
Model performance and generalizability were summarized in Table \ref{ModelPerformance} for each approach, measured in terms of training and validation loss. The first approach, BLS without BRS, exhibited the poorest performance, with training and validation losses of 0.0962 ± 0.0012 and 0.0945 ± 0.0020, respectively. Similarly, BLS with BRS (WV) showed comparable poor results, with losses of 0.0948 ± 0.0003 and 0.0924 ± 0.0014. In contrast, BLS with OV achieved the best performance, with significantly lower training and validation losses of 0.0339 ± 0.0062 and 0.0389 ± 0.0057, respectively. The third approach, BLS with SLS, delivered intermediate results, with training and validation losses of 0.0462 ± 0.0041 and 0.0442 ± 0.0034, respectively.

\begin{table}[ht]
\centering
\caption{\centering Training and validation hybrid loss for various approaches for 5-fold cross-validation.}\label{ModelPerformance}
\begin{tabular}{lcc}
\toprule
\textbf{Approach} & \textbf{Training Loss} & \textbf{Validation Loss} \\
\midrule
BLS without BRS (WV) &  0.0962 $\pm$ 0.0012  & 0.0945 $\pm$ 0.0020 \\
BLS with BRS (WV)    & 0.0948 $\pm$ 0.0003 & 0.0924 $\pm$ 0.0014 \\
BLS with BRS (SLS)    & 0.0462 $\pm$ 0.0041 & 0.0442 $\pm$ 0.0034 \\
BLS with BRS (OV)     & 0.0339 $\pm$ 0.0062 & 0.0389 $\pm$ 0.0057 \\
\bottomrule
\end{tabular}
\label{tab:dice_losses}
\end{table}

\begin{figure}[h!]
\centering
\includegraphics[width=1.4\textwidth, trim=10 250 150 220, clip]{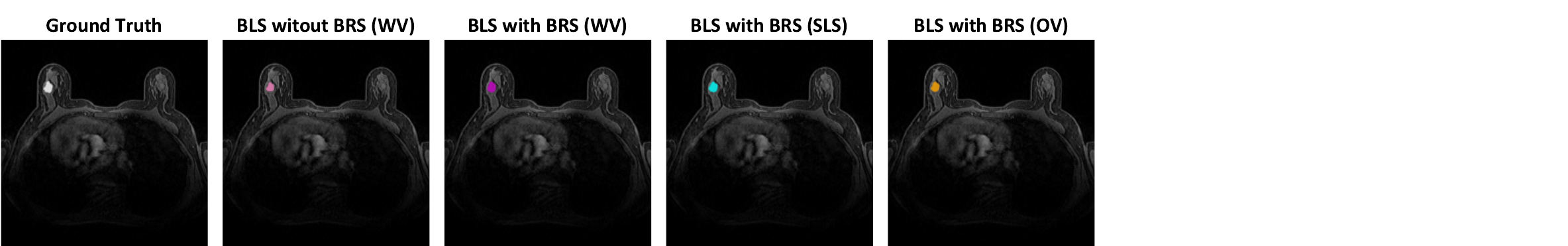}
\caption{\centering  Prediction results on test dataset for different methods } \label{LesionPrediction}
\end{figure}

Figure \ref{LesionPrediction} depicts prediction results on one of the test slices for all approaches along with ground truth mask. 

\subsection{Lesion Segmentation}\label{sec3}
Table \ref{Evaluation metrics} demonstrates an overview of the evaluation metrics, including the average Dice, IoU, precision, and recall scores for the test dataset. As shown in the table, the best performance belongs to BLS with BRS (OV), achieving the highest scores across all metrics: 0.640 in Dice, 0.539 in IoU, 0.705 in precision, and 0.640 in recall. In contrast, BLS without BRS had the lowest performance, with scores of 0.414 in Dice, 0.328 in IoU, 0.586 in precision, and 0.354 in recall. The second-best results were obtained by BLS with BRS (SLS), which achieved 0.542 in Dice, 0.447 in IoU, 0.542 in precision, and 0.605 in recall.

\begin{table}[ht]
\centering
\caption{\centering Evaluation metrics for different approaches}\label{Evaluation metrics}
\begin{tabular}{lcccc}
\toprule
\textbf{Methods} & \textbf{Dice}$_{\text{avg}}$ & \textbf{IoU}$_{\text{avg}}$ & \textbf{Precision}$_{\text{avg}}$ & \textbf{Recall}$_{\text{avg}}$\\
\midrule
BLS without BRS (WV)  & 0.414 & 0.328 & 0.586 & 0.354   \\
BLS with BRS (WV)     & 0.423 & 0.336 & 0.461 & 0.482 \\
BLS with BRS (SLS)    & 0.542 & 0.447 & 0.542 & 0.605 \\
BLS with BRS (OV)     & \textbf{0.640} & \textbf{0.539} & \textbf{0.705} & \textbf{0.640} \\
\bottomrule
\end{tabular}
\label{tab:dice_losses}
\end{table}

To assess the number of false positives and false negatives in the model’s predictions for threshold 0.5 on the test dataset, a thorough evaluation was conducted. Table \ref{FPFN} presents the number of false positive and false negative volumes across different lesion sizes for each method. Notably, BLS with BRS (OV) had the lowest number of misclassified lesions larger than 20 mm$^3$, with only 35 false positives, and the fewest undetected lesions in the test dataset, with just one false negative. BLS without BRS (WV), showed similar performance compared to BLS with BRS but had a higher number of false positives for lesions smaller than 10 mm$^3$, totaling 20, and performed worse in detecting lesions larger than 20 mm$^3$, with five false negatives. The other methods fell between these two approaches but exhibited overall poorer results. Furthermore, the largest false positives (larger than 20 mm$^3$) for all approaches were incorrectly identified in the early and late slices, where artifacts were present, and the breast region was not fully developed.

\begin{table}[ht]
\centering
\caption{\centering Number of FP and FN lesion volume (mm\textsuperscript{3})  for threshold= 0.5}
\label{FPFN}
\begin{tabular}{@{}lcccccc@{}}
\toprule
\textbf{Method}  & \multicolumn{3}{c}{\textbf{False Positives}} & \multicolumn{3}{c}{\textbf{False Negatives}} \\ 
\cmidrule(lr){2-4} \cmidrule(lr){5-7} 
                & {\tiny V < 10 }   & {\tiny 10 < V < 20} & {\tiny V > 20} & {\tiny V < 10} &  {\tiny 10 < V < 20} & {\tiny V > 20} \\ 
\midrule
BLS without BRS (WV)  & 20  & 5  & 35  & 0  & 2  & 5  \\
BLS with BRS (WV)     & 39  & 12 & 45  & 0  & 1  & 0  \\
BLS with BRS (SLS)    & 25  & 11 & 46  & 1  & 1  & 2  \\
BLS with BRS (OV)     & 11  & 9  & 35  & 1  & 2  & 1  \\ 
\bottomrule
\end{tabular}
\end{table}

\begin{figure}[h!]
\centering
\includegraphics[width=1.05\textwidth, trim=0 55 0 0, clip]{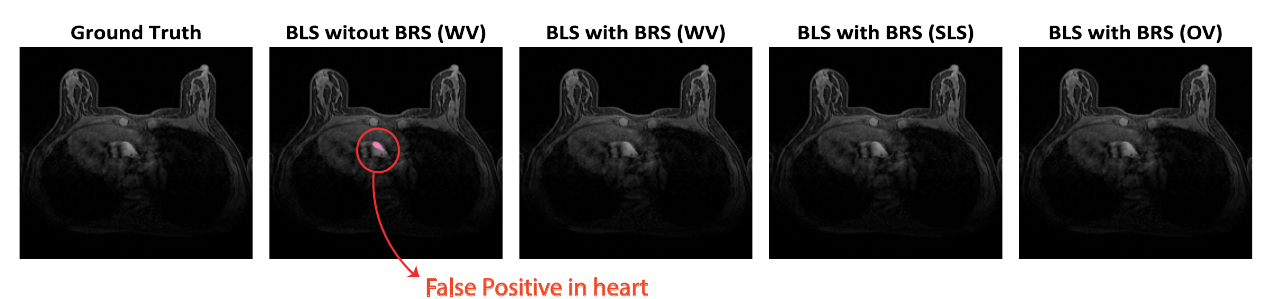}
\caption{\centering  Example of False Positive on test dataset for different methods } \label{HeartFP}
\end{figure}

Figure \ref{HeartFP} illustrates an example of a false positive in a high-intensity region for BLS without BRS (WV). The predicted pixels, incorrectly classified as a lesion, are highlighted with a circle. In this case, the heart was mistakenly identified as a lesion, emphasizing the importance of removing noisy and high-intensity areas to reduce misclassification errors.

\subsection{Carbon Footprint}\label{sec3}
Carbon footprint has been a challenging subject in the last decades and more relevant nowadays in AI applications. Since training of large DL models needs high demand of electricity and therefore resulting in carbon emission originating from energy production, dealing with carbon footprint is a vital step in deep learning applications. Table \ref{training time} demonstrates results on training time during each approach showing that BLS with BRS (OV) approach has the least training time by 16 $\pm$ 4 min per fold and less epochs to get the best performance. On the other hand, BLS without BRS (WV) has the longest training time among all with 75 $\pm$ 21 min per each fold.

\begin{table}[ht]
\centering
\caption{\centering Training time, best and last epochs across different approaches.}\label{training time}
\begin{tabular}{lccc}
\toprule
\textbf{Approach} & \textbf{TT per fold (min)} & \textbf{CFP per fold}  & \textbf{Last epochs}\\
\midrule
BLS without BRS (WV) &  75 $\pm$ 21  & 0.59 $\pm$ 0.17 & 30 $\pm$ 9 \\
BLS with BRS (WV)   & 68 $\pm$ 13 & 0.54 $\pm$ 0.10 & 27 $\pm$ 5\\
BLS with BRS (SLS)    & 25 $\pm$ 12 & 0.20 $\pm$ 0.10 & 29 $\pm$ 14\\
BLS with BRS (OV)     & 16 $\pm$ 4 & 0.13 $\pm$ 0.03 & 24 $\pm$ 7\\

\bottomrule
\end{tabular}
\label{tab:training_inference_times}
\end{table}

\section{Discussion}\label{sec4}
This study evaluates four different approaches for BLS, comparing one without BRS and three using BRS. The results indicate that model performance varies significantly across validation and test datasets, particularly in how shape modifications influence training.

As shown in Table \ref{InputShape}, the NIFTI image shape is substantially affected by data analysis approach. The most notable change occurs in the last two approaches, where dimensional modifications extend beyond 2D slices to the entire volume. In the OV approach, the data is more balanced, with 2D slices resized by nearly 46\%, which is a major contrast to the other methods. This significant transformation is clearly visible in Figure \ref{Datashape}.

In terms of performance, Table \ref{ModelPerformance} highlights a clear advantage for the OV approach over the others. Not only OV exhibit lower training loss, but it also maintains superior validation performance. On the other hand, the WV approaches, do not show a substantial impact from breast region segmentation. This suggests that eliminating noisy areas in the whole volume does not dramatically alter model performance. One possible explanation is that over 70\% of slices in the WV approach contain no annotated lesions, which biases the model toward negative samples. In contrast, the selective inclusion of slices in OV and SLS provides a more balanced learning process, improving efficiency and overall segmentation quality.

A deeper analysis of lesion distribution, shown in Figure \ref{OverlayMap}, further illustrates the key differences between approaches. The lesion distribution for WV (both with and without BRS) is closely similar to that of SLS, as negative slices contain no lesion pixels. However, the breast region map, presented in Figure \ref{BRS16Color}, shows distinct variations. In the SLS approach, the breast region concentration is lower, leading to a reduced \( H_{\text{max} \vert \text{mid}} \)  value compared to WV-based methods. Meanwhile, the OV approach presents the highest concentration of lesion pixels along the x-axis, resulting in a unique intensity histogram. Notably, lesion distribution along the y-axis remains consistent across all approaches, showing a higher lesion density in the left breast compared to the right.

A broader comparison of breast region maps, as depicted in Figure \ref{BRS16Color}, reveals that approaches other than OV exhibit greater variability (higher SD) due to a less concentrated breast region distribution. The histogram intensity for the x-axis (at $y_{\text{mid}}$) confirms that whole-volume approaches have the highest \( H_{\text{max} \vert \text{mid}} \)  values, while OV shows a more localized region distribution. This suggests that OV provides a more stable and concentrated dataset, improving segmentation accuracy.

The effectiveness of BRS is further reinforced in Table \ref{Evaluation metrics}, which demonstrates that BLS with BRS yields the best results, particularly in the OV approach. Additionally, Table \ref{FPFN} supports this conclusion by analyzing FP and FN. The most significant FP (greater than 20 mm$^3$) were mistakenly detected in the early and late slices due to artifacts, where the breast region was not fully formed.

Finally from a computational efficiency perspective, optimizing the volume in OV led to a dramatic reduction in training time and carbon footprint. Compared to WV-based approaches, the optimization reduced energy consumption by approximately 450\%, highlighting the unnecessary computational cost of processing non-informative regions. Furthermore, the OV approach required fewer epochs to reach convergence, demonstrating that a well-structured dataset enhances learning efficiency.

Future studies should explore whether updating SD for new datasets would further improve segmentation outcomes. Additionally, while increasing data diversity from multiple centers may enhance model generalization, lesion annotation remains a significant challenge requiring expert intervention. Although this study focuses on 2D approaches, future research could extend the analysis to 2.5D and 3D methodologies to evaluate their relative efficiency in the OV approach.

\section{Conclusion}\label{sec5}
In this study, a 2D deep learning model, UNet++, was developed to assess the impact of breast region segmentation (BRS) on the performance of breast lesion segmentation (BLS). The model was separately trained with four different data shapes, with each shape representing a distinct region of interest, to evaluate their impact on training efficiency. Identical hyperparameters were used across all approaches, and the learning rate was self-adjusted during training. A 5-fold cross-validation method, combined with a hybrid loss function, was employed to optimize the model's learnable parameters.

The results revealed that incorporating BRS on the entire volume slightly improved the model's performance. More significantly, data analysis from BRS greatly enhanced training efficiency. Among the different BRS approaches, segmentation based on the whole volume (WV) yielded the lowest performance, while segmentation based on the optimized volume (OV) achieved the best results. Additionally, training time was significantly reduced when using the OV approach, which also resulted in the smallest carbon footprint.

In conclusion, BLS using BRS with OV not only provided the best performance but also demonstrated the most environmentally sustainable model, with the least carbon footprint. This approach offers a green solution for training on large datasets while achieving superior results, highlighting the importance of sustainability in deep learning applications.

\section*{Declarations}

\subsection*{Ethics compliance}
All patients enrolled in the IMAGINE project cohorts received approval from the regional ethics committee. All methods were conducted in accordance with relevant guidelines and regulations, and informed consent was obtained from all participants.

\subsection*{Data availability}
The Stavanger dataset analyzed in this study contains sensitive patient information and therefore not publicly available. It will be available upon reasonable request by contacting Endre Grøvik through the institution.

\subsection*{Code availability}
The code for data preprocessing, data analyzing, visualization, modeling and evaluation is available on GitHub. To access the code repository, please follow the link on \href{https://github.com/SamNarimani-lab/Breast.git}{GitHub.}

\subsection*{Acknowledgment}
We would like to express our heartfelt gratitude to More and Romsdal Hospital Trust for their support, as well as to Stavanger University Hospital for contributing resources and data. We are particularly grateful to Kathrine Røe Redalen and Marianne Høybakk Almås for their dedicated assistance during the research process. This project was made possible through funding from the Central Norway Regional Health Authority, under the IMAGINE project.

\subsection*{Author contributions}
\textbf{Sam Narimani}: Drafted the introduction, methodology, preprocessing and data insight, and authored the results, discussion, and conclusion sections as well as conducting and testing programming code. \\
\textbf{Solveig Roth Hoff}: Contributed to the writing of the introduction, major revision of all sections, Second expert radiologist for annotation of test set.\\ 
\textbf{Kathinka Dæhli Kurz}: Responsible for data acquisition and preparation, annotation of breast lesions , revising the manuscript . \\
\textbf{Kjell-Inge Gjesdal}: Set up MRI protocols. \\
\textbf{Jürgen Geisler}: Writing introduction and revision of the draft.\\
\textbf{Endre Grøvik}: Supervised the project, acquired funding and draft revision .



\end{document}